\documentclass[final]{cvpr}

\usepackage{times}
\usepackage{epsfig}
\usepackage{graphicx}
\usepackage{amsmath}
\usepackage{amssymb}
\usepackage{multirow}
\usepackage{booktabs}
\usepackage{amsfonts,amssymb}

\usepackage{amsmath,amsfonts,bm}

\def\eqref#1{equation~\ref{#1}}

\def\1{\bm{1}}

\DeclareMathAlphabet{\mathsfit}{\encodingdefault}{\sfdefault}{m}{sl}
\SetMathAlphabet{\mathsfit}{bold}{\encodingdefault}{\sfdefault}{bx}{n}

\DeclareMathOperator*{\argmin}{arg\,min}

\usepackage{subcaption}
\usepackage[pagebackref=true,breaklinks=true,colorlinks]{hyperref}

\usepackage{algorithm}
\usepackage{algorithmic}
\usepackage{soul}
\usepackage{xcolor}

\usepackage{color}
\newcommand{\xw}[1]{\textcolor{blue}{#1}}
\newcommand{\xwc}[1]{\textcolor{red}{\bf [Comments: #1] }}

\newcommand{\bo}[1]{\textcolor{black}{#1}}

\newcommand{\name}{{CAFE}}

\newcommand{\bx}{\bm{x}}

\newcommand{\bs}{\bm{s}}

\newcommand{\bff}{\bm{f}}
\newcommand{\bF}{\bm{F}}
\newcommand{\btheta}{\bm{\theta}}

\newcommand{\mt}{\mathcal{T}}
\newcommand{\ms}{\mathcal{S}}

\def\cvprPaperID{5660} 
\def\confYear{CVPR 2022}

\begin{document}

\title{CAFE: Learning to Condense Dataset by Aligning Features}

\author{
Kai Wang\textsuperscript{1}\thanks{Equal contribution. (kai.wang@comp.nus.edu.sg, bo.zhao@ed.ac.uk)} 
\quad Bo Zhao\textsuperscript{2}\footnotemark[1]
\quad Xiangyu Peng\textsuperscript{1}
\quad Zheng Zhu\textsuperscript{3}
\quad Shuo Yang\textsuperscript{4}
\quad Shuo Wang\textsuperscript{5}
\\
\quad Guan Huang \textsuperscript{3}
\quad Hakan Bilen\textsuperscript{2}
\quad Xinchao Wang \textsuperscript{1}
\quad Yang You\textsuperscript{1}\thanks{Corresponding author (youy@comp.nus.edu.sg).}
\\
\textsuperscript{1}{National University of Singapore}
\quad \textsuperscript{2}{The University of Edinburgh}
\quad \textsuperscript{3}{PhiGent Robotics}
\\
\quad \textsuperscript{4}{University of Technology Sydney}
\quad \textsuperscript{5}{Institute of Automation, Chinese Academy of Sciences}

\\
\small{Code: \url{https://github.com/kaiwang960112/CAFE}}
}


\maketitle

\begin{abstract}
Dataset condensation aims at reducing the network 
training effort through condensing a 
cumbersome training set into a compact 
synthetic one. State-of-the-art approaches 
largely rely on learning the synthetic data 
by matching the gradients between the real
and synthetic data batches. Despite the 
intuitive motivation and promising results,
such gradient-based methods, by nature, 
easily over-fit to a biased set of samples
that produce dominant gradients,
and thus lack a global supervision of data distribution. 
In this paper, we propose a novel scheme 
to Condense dataset by Aligning FEatures (CAFE),
which explicitly attempts to preserve 
the real-feature distribution as well
as the discriminant power of the resulting 
synthetic set, lending itself to strong 
generalization capability to various architectures.
At the heart of our approach is an effective strategy
to align features from the real and synthetic data 
across various scales, while accounting 
for the classification of real samples. 
Our scheme is further backed up by a novel 
dynamic bi-level optimization, which adaptively 
adjusts parameter updates to prevent over-/under-fitting.
We validate the proposed CAFE across various datasets,
and demonstrate that it generally outperforms the state
of the art: on the SVHN dataset, for example, 
the performance gain is up to 11\%. Extensive 
experiments and analysis verify the effectiveness
and necessity of proposed designs.
\end{abstract}

\section{Introduction}
Deep neural networks~(DNNs) have demonstrated 
unprecedented results in many if not all
applications 
in computer vision~\cite{deng2009imagenet, OpenImages, wang2021efficient, peng2020suppressing, pascal-voc-2012, lin2014microsoft, real2017youtube, damen2020epic,yang2021single, zhang2021learning, peng2022crafting, wang2021mask, wang2020region, wang2020suppressing}.
These gratifying results, 
nevertheless, come with costs:
the training of DNNs
heavily rely on the
sheer amount of data, sometimes up to tens of millions of samples,
which consequently requires
enormous computational resources.

\begin{figure}[t]
\centering
\subfloat[\bo{The gradient distribution changes from \bo{a uniform} to long-tailed distribution during the training. Meanwhile, the overlap of large-gradient samples are small among different architectures.}]{\includegraphics[width = 1.0\linewidth]{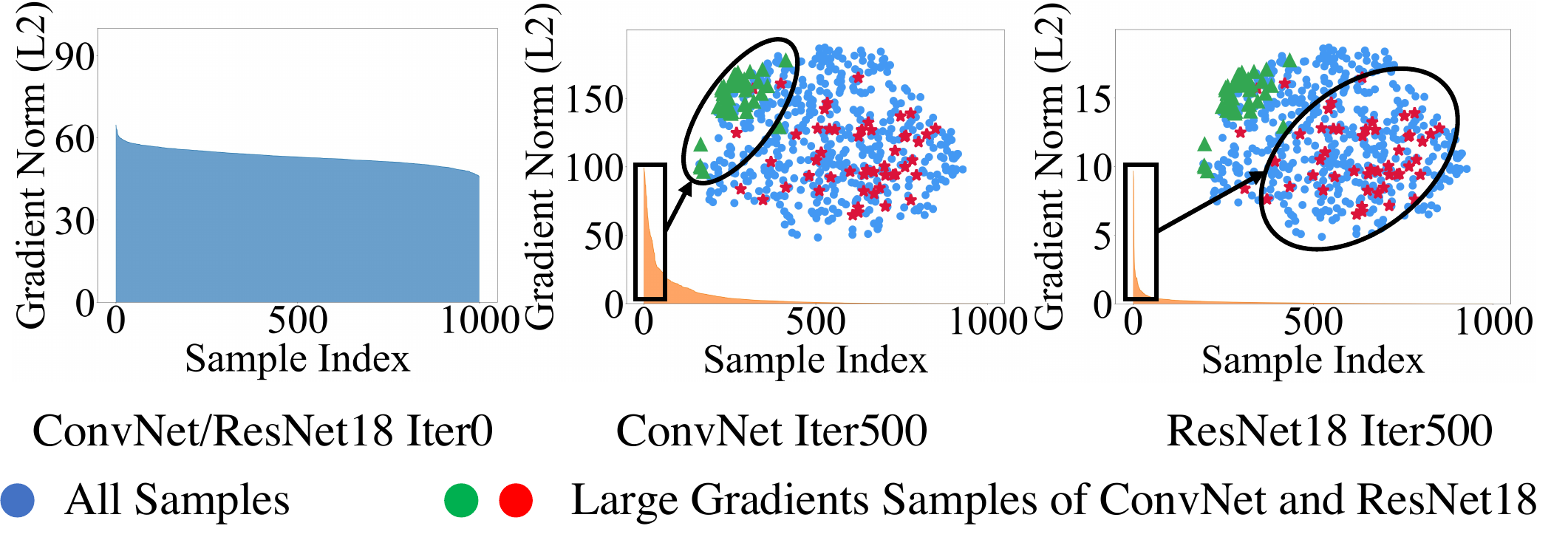}\label{fig:motivation_zero}}\hfill
\subfloat[\bo{The visualization of synthetic images and their distributions generated by \textbf{gradient matching} and \textbf{CAFE}. ConvNet is used.}]{\includegraphics[width = 1.0\linewidth]{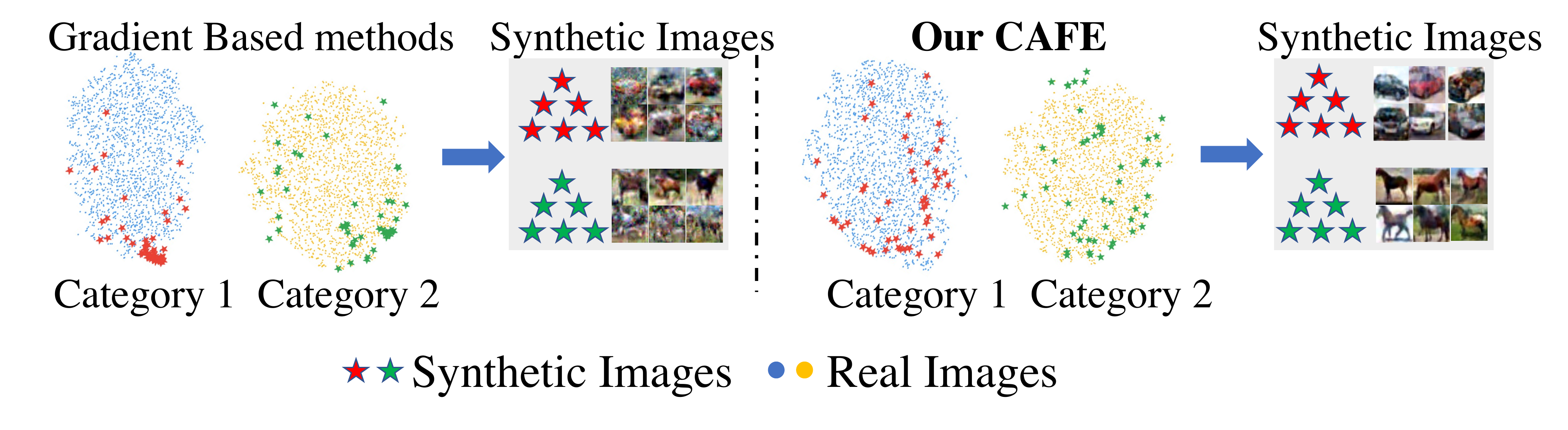}\label{fig:motivation_ab}}\hfill
\caption{(a) At the later training stage, most examples do not contribute meaningful gradients, making the synthetic set learned by gradient matching extremely bias towards those large-gradient samples, which downgrades its generalization to unseen architectures. (b) Compared with gradient-based method~\cite{zhao2021DC}, the synthetic set learned by our approach effectively captures the whole distribution thus generalizes well to other network architectures.}

\label{fig:motivation}
\end{figure}

Numerous research endeavours have, therefore,
focused on alleviating the 
cumbersome training process
through constructing small 
training sets~\cite{agarwal2004approximating, feldman2007ptas, chen2010super, feldman2011scalable, wolf2011kcenter, sener2017active, sinha2020small,yang2021free,yang2021bridge,zhang2022fewshot}. 
One classic approach is known as 
coreset or subset selection~\cite{agarwal2004approximating, sener2017active, feldman2020introduction},
which aims to obtain a subset of salient data points
to represent the original dataset of interest.
Nevertheless, 
coreset selection is typically a NP-hard problem~\cite{knoblauch2020optimal},
making it computationally intractable
over large-scale datasets.
Most existing approaches have thus resorted
to greedy algorithms with 
heuristics~\cite{chen2010super, sener2017active,aljundi2019gradient,yoon2021online,toneva2019empirical}
to speed up the process
by trading-off the optimality.

Recently, dataset condensation~\cite{wang2018dataset, zhao2021DC, cazenavette2022dataset}
has emerged as a competent alternative 
with promising results.
The goal of dataset condensation is,
as its name indicates, to 
condense a large training set into a small synthetic one,
upon which DNNs are trained
and expected to preserve the performances.
Along this line, the pioneering approach of~\cite{wang2018dataset} 
proposed a meta-learning-based 
strategy; however, the nest-loop optimization
precludes its scaling up to large-scale
in-the-wild datasets.
The work of~\cite{zhao2021DC}  alleviates this issue by enforcing 
the batch gradients of the synthetic samples to approach those of 
the original ones, which bypasses the recursive 
computations and achieves impressive results. 
The optimization of the synthetic examples is explicitly 
supervised by minimizing the distance between 
the gradients produced by the synthetic dataset and the real dataset.

However, gradient matching method has two potential problems. 
First, due to the memorization effect 
of deep neural  networks~\cite{zhang2017understanding}, 
only a small number of hard examples or noises produce dominant gradients over the network parameters. Thus, gradient matching may overlook those representative but easy samples, while overfit to those hard samples or noises. 
Second, these hard examples that produce large gradients 
may vary across different architectures;
relying solely on gradients, therefore,
will yield poor generalization performance to unseen architectures.
The distributions of gradients and hard examples are 
illustrated in Fig.~\ref{fig:motivation_zero}.
The synthetic data learned by gradient matching may be highly biased towards a small number of unrepresentative data points, which is illustrated in Fig. \ref{fig:motivation_ab}.

To go beyond the learning bias and better capture the whole dataset distribution, 
in this paper, we propose
a novel strategy 
to  Condense dataset by Aligning FEatures, 
termed as \name.
Unlike the approach of~\cite{zhao2021DC},
we account for the distribution
consistency between synthetic and real datasets by
applying distribution-level supervision.
Our approach, through matching the features that involve all intermediary layers, 
expands the attention across all samples and hence
provides a much more comprehensive 
characterization of the distribution while 
avoiding over-fitting on hard or noisy samples.
Such distribution-level supervision will,
in turn, endow {\name} with 
stronger generalization power
than gradient-based methods,
since the hard examples may easily 
vary across different architectures. 


Specifically, we impose two
complementary losses into the objective
of {\name}. The first one 
concerns capturing the data distribution, in which 
the layer-wise alignment
between the features of 
the real and synthetic samples
is enforced
and further the distribution is preserved.
The second loss,
on the other hand,
concerns discrimination.
Intuitively, 
the learned synthetic samples
from one class 
should well represent the
corresponding clusters 
of the real samples.
Hence, we may treat each
real sample as a testing sample,
and classify it based on its affinity
to the synthetic clusters.
Our second loss is then defined upon
the classification result of the real samples,
which, effectively, 
injects the discriminant capabilities into
the synthetic samples.

The proposed {\name} is further backed up by
a novel bi-level optimization scheme,
which allows our network and synthetic data
to be updated through a  customized number of SGD steps.
Such a dynamic optimization strategy,
in practice, largely alleviates the under-
and over-fitting issues of prior methods.
We conduct experiments on several 
popular benchmarks and demonstrate that,
the results yielded by {\name} are significantly superior to
the state of the art:
on the SVHN dataset, for example,
our method outperforms the runner-up by $11\%$ when learning 1 image/class synthetic set.
We also especially prove that synthetic set learned by 
our method has better generalization ability than that learned by \cite{zhao2021DC}.

In summary, our contribution is a
novel and effective approach for condensing
datasets, achieved through aligning layer-wise features
between the real and synthetic data,
and meanwhile explicitly encoding the discriminant power
into the synthetic clusters. 
In addition, a new bi-level optimization scheme
is introduced,
so as to adaptively
alter the number of SGD steps.
These strategies jointly enable
the proposed {\name} to
well characterize the distribution
of the original samples,
yielding state-of-the-art performances
with strong generalization and robustness
across various learning settings.

\section{Related Work}
\paragraph{Dataset Condensation.}
Several methods have been proposed to improve the performance, scalability and efficiency of dataset condensation. Based on the meta-learning method proposed in \cite{wang2018dataset}, some works \cite{bohdal2020flexible, nguyen2021dataset, nguyen2021datasetnips} try to simplify the inner-loop optimization of a classification model by training with ridge regression which has a closed-form solution. \cite{such2020generative} trains a generative network to produce the synthetic set. The generative network is trained using the same objective as \cite{wang2018dataset}.
To improve the data efficiency of \cite{zhao2021DC}, differentiable Siamese augmentation is proposed in \cite{zhao2021DSA}. They enable the synthetic data to train neural networks with data augmentation effectively. 

A recent work \cite{zhao2021DM} also learns synthetic set with feature distribution matching. Our method is different from it in three main aspects: 1) we match layer-wise features while \cite{zhao2021DM} only uses final-layer features; 2) we further explicitly enable the synthetic images to be discriminative as the classifier (\ie Sec. \ref{sec:discrimination}); 3) our method includes a dynamic bi-level optimization which can boost the performance with adaptive SGD steps, while \cite{zhao2021DM} tries to reduce the training cost by dropping the bi-level optimization.


\paragraph{Coreset Selection.}
The classic technique to condense the training set size is coreset or subset selection \cite{agarwal2004approximating, chen2010super, feldman2013turning, wei2015submodularity}. Most of these methods incrementally select important data points based on heuristic selection criteria. For example, \cite{sener2017active} selects data points that can approach the cluster centers.  \cite{aljundi2019gradient} tries to maximize the diversity of samples in the gradient space. \cite{toneva2019empirical} measures the forgetfulness of trained samples during network training and drops those that are not easy to forget. However, these heuristic selection criteria cannot ensure that the selected subset is optimal for training models, especially for deep neural networks. In addition, greedy sample selection algorithms are unable to guarantee that the selected subset is optimal to satisfy the criterion.  

\paragraph{Generative Models}
Our work is also closely related to generative model such as auto-encoder \cite{kingma2013auto} and generative adversarial networks (GANs) \cite{goodfellow2014generative, mirza2014conditional}. The difference is that image generation aims to synthesize real-looking images that can fool human beings, while our goal is to generate informative training samples that can be used to train deep neural networks more efficiently. As shown in \cite{zhao2021DC}, concerning  training models, the efficiency of these images generated by GANs closes to that of randomly sampled real images. In contrast, our method can synthesize better training images that significantly outperform those selected real images in terms of model training.


\begin{figure*}[htp]
\vspace{-5pt}
\centering
\includegraphics[width=0.9\textwidth]{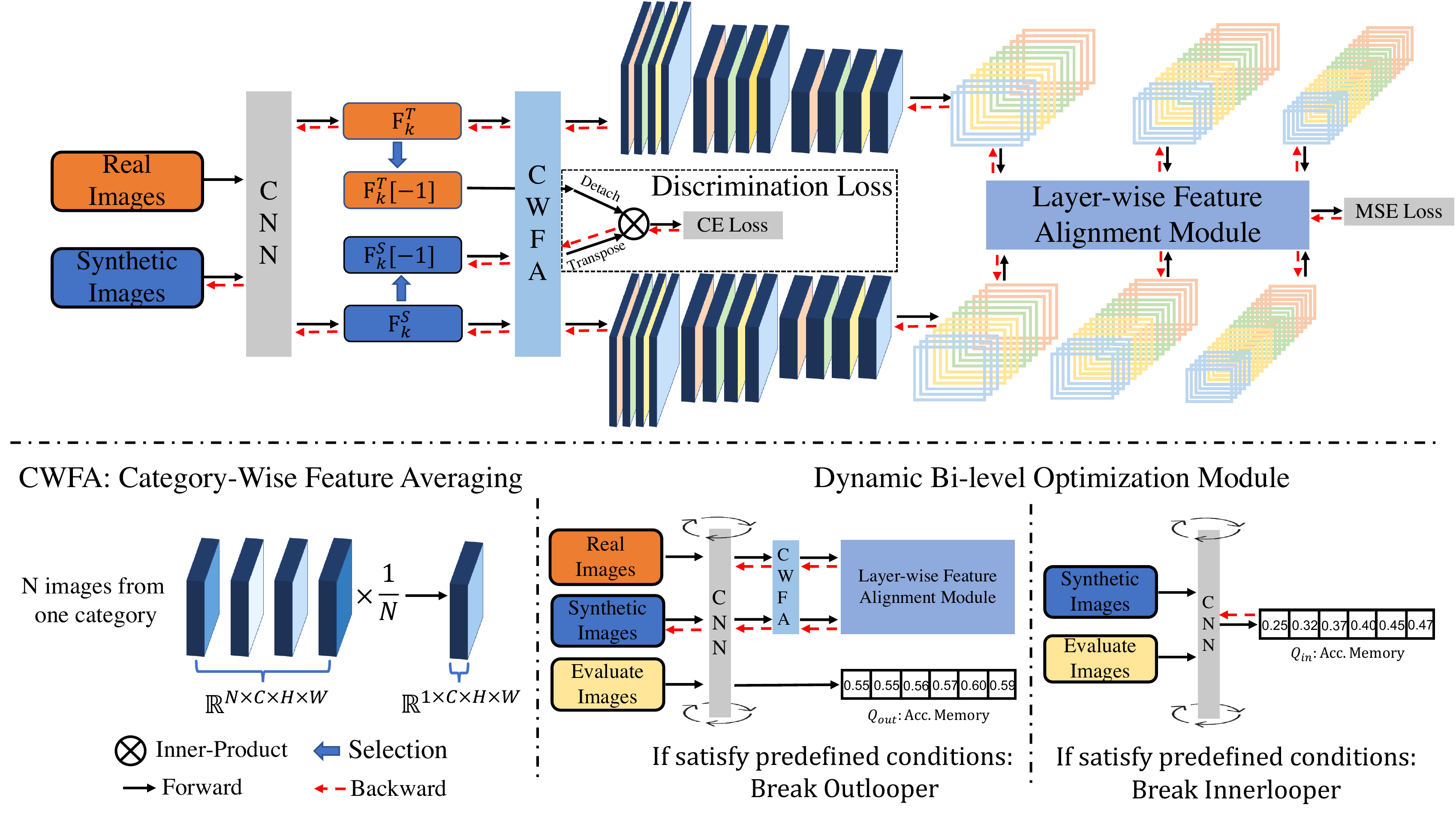}
\vspace{-5pt}
\caption{Illustration of the  proposed CAFE method. The CAFE consists of a layer-wise feature alignment module to capture the accurate distribution of the original large-scale dataset, a discrimination loss for mining the discriminate samples from real dataset, and a dynamic bi-level optimization module to reduce the influence of under- and over-fitting on synthetic images.}

\label{pipeline}
\end{figure*}

\section{Method}
In this section, we first briefly overview the proposed \name. Then, we introduce three carefully designed modules: layer-wise feature alignment module,
discrimination loss,
and dynamic bi-level optimization module. 


\subsection{Overview}
Dataset condensation aims to condense a large-scale dataset $\mt = \{(\bx_i, y_i)\}|^{|\mt|}_{i=1}$ into small (synthetic) dataset $\ms = \{(\bs_j, y_j)\}|^{|\ms|}_{j=1}$ while achieving similar generalization performance. Fig. \ref{pipeline} illustrates the proposed method. First, we sample two data batches from the large-scale dataset $\mt$ and the learnable synthetic set $\ms$ respectively, and then extract the features using neural network $\phi_{\btheta}(\cdot)$ which is parameterized with $\btheta$. 
To capture the distribution of $\mt$ accurately, layer-wise feature alignment module is designed, in which we minimize the difference of layer-wise feature maps of real and synthetic images using Mean Square Error (MSE). To enable learning discriminative synthetic images, we use the feature centers of synthetic images of each class to classify the real images by computing their inner-product and cross-entropy loss. The synthetic images are updated by minimizing the above two losses, which is the outer-loop. Then, we update the network $\phi_{\btheta}(\cdot)$ by minimizing the cross-entropy loss on synthetic images, which is the inner-loop. The synthetic images and network are alternatively using a  novel dynamic bi-level optimization algorithm which avoids the over- or under-fitting on synthetic dataset and breaks the outer- and inner-loop automatically.

\subsection{Layer-wise Features Alignment}
\label{dcfm}

As mentioned above, previous works \cite{zhao2021DC,zhao2021DSA} compare the differences of gradients between real and synthetic data. Such objective produces samples with large gradients, but these samples fail to capture the distribution of the original dataset (illustrated in Fig. \ref{fig:distribution}). Thus, it may have poor performance when generalizing to unseen architectures. To tackle this issue, we design Category-Wise Feature Averaging (CWFA), illustrated in Fig. \ref{pipeline}, to measure the feature difference between $\mt$ and $\ms$ at each convolutional layer. 
Specifically, we sample a batch of real data $\mt_k$ and synthetic data $\ms_k$ with the same label $k$ and the batch size $N$ and $M$, from $\mt$ and $\ms$ respectively. 
We embed each real and synthetic datum using network $\phi_{\btheta}(\cdot)$ with $L$ layers (except the output layer) and obtain the layer-wise features $\bF^\mt_k = [\bff^\mt_{k, 1}, \bff^\mt_{k, 2}, ..., \bff^\mt_{k, L}] = \phi_{\btheta}(\mt_k)$ and $\bF^\ms_k = [\bff^\ms_{k, 1}, \bff^\ms_{k, 2}, ..., \bff^\ms_{k, L}] = \phi_{\btheta}(\ms_k)$. The $l^\text{th}$ layer feature $\bff^\mt_{k, l} \in \mathbb{R}^{N \times C'}$ is reduced to ${\bar{\bff}^\mt_{k, l}} \in \mathbb{R}^{1 \times C'}$ by averaging the $N$ samples in the real data batch, where $C' = C \times H \times W$ and it refers to the feature size of corresponding layer. Similarly, we obtain ${\bar{\bff}^\ms_{k, l}}$ for synthetic data batch.

Then, MSE is applied to calculate the feature distribution matching loss $\mathcal{L}_\text{f}$ for every layer, which is formulated as
\begin{equation}
    \mathcal{L}_{\text{f}} =  \sum_{k=1}^{K}\sum_{l=1}^{L} |{\bar{\bff}^\ms_{k, l}} - {\bar{\bff}^\mt_{k, l}}|^2,
\end{equation}
where $K$ is the number of categories in a dataset. 

\subsection{Discrimination Loss}
\label{sec:discrimination}
Though the layer-wise feature alignment can capture the distribution of original dataset, it may overlook the discriminative sample mining. We hold the view that an informative synthetic set could be used as a classifier to classify real samples. 
Based on this, we calculate the classification loss in the last-layer feature space.
we obtain the synthetic feature center $\bar{\bff}^\ms_{k, L} \in \mathbb{R}^{1 \times C'}$ of each category $k$ by averaging the batch. We concatenate the feature centers $\bar{\bF}^\ms_{L} = [\bar{\bff}^\ms_{1, L}, \bar{\bff}^\ms_{2, L}, ..., \bar{\bff}^\ms_{K, L}]$ and also real data $\bF^\mt_{L} = [\bff^\mt_{1, L}, \bff^\mt_{2, L}, ..., \bff^\mt_{K, L}]$ from all classes.
The real data is classified using the inner-product between real data and the synthetic centers 
\begin{equation}
\mathbf{O} = \left< \bF^\mt_{L}, {(\bar{\bF}^\ms_{L})}^{\text{T}} \right>,
\end{equation}
where $\mathbf{O} \in \mathbb{R}^{N' \times K}$ contains the logists of $N' = K \times N$ real data points. The classification loss is 
\begin{equation}
    \mathcal{L}_{\text{d}} = -\frac{1}{N'}\Sigma_{i=1}^{N'}\log p_i,
\end{equation}
where the probability $p_i$ is the softmax value corresponding to its ground-truth label over all classes $p_i = \text{softmax}(\mathbf{O}_i)$. The total loss for learning synthetic images is 
\begin{equation}
\mathcal{L}_{\text{total}} =  \mathcal{L}_{\text{f}} + \beta\mathcal{L}_{\text{d}},
\label{eq.loss_img}
\end{equation}
where $\beta$ is a positive scalar weight of $\mathcal{L}_{\text{d}}$. We study the influence of $\beta$ in Sec. \ref{abl_stdy}.
The synthetic set is updated by minimizing $\mathcal{L}_{\text{total}}$:
\begin{equation}
\ms \leftarrow \argmin_\ms \mathcal{L}_{\text{total}}
\label{eq.updateS}
\end{equation}

\begin{algorithm}[h]
	\caption {Dynamic Bi-level Optimization}{$\mt$ and $\ms$ are the real the synthetic datasets. $\epsilon$ is a random sampling function for selecting $N$ images from $K$ categories. $Q_{out}$ and $Q_{in}$ are the queues to save the performance on real dataset in outer- and inner-loop, respectively. div($\cdot$) is a function that calculates the difference between maximum and minimum values of $Q_{out}$ and $Q_{in}$. $\gamma$ is the maximum length of queues. The default loop numbers of DC are $l_{out}$ and $l_{in}$. $l_{c}^\cdot$ represents the loop number of CAFE.}

	\begin{algorithmic}[1]
	\small
 		    \WHILE{not converged} 
            \STATE randomly initialize $\btheta$, $Q_{out}$ = []; $Q_{in}$ = []; $l_{c}^{out}$=$l_{c}^{in}$ = 0.
    		\WHILE{True} 
            \STATE updating $\ms$ using Eq. \ref{eq.updateS};  $l_{c}^{out}$ += 1. \algorithmiccomment{outer-loop}
            \STATE acc. = get$\_$acc$(\epsilon(K, N))$; $Q_{out}$.append(acc.).
            \IF {~$|Q_{out}|$ == $\gamma$ \textbf{and} div($Q_{out}$)$<\lambda_1$ \textbf{or} $l_{c}^{out}>l_{out}$ ~}
            \STATE $l_{c}^{out}$ = 0, $Q_{out}$ = [].
    		\STATE \textbf{Break}
    		\ELSE \STATE $Q_{out}$.pop[0].
    		\ENDIF
    		\WHILE{True}
            \STATE updating $\btheta$ using Eq.  \ref{eq.updateTheta}. \algorithmiccomment{inner-loop}
            \STATE acc. = get$\_$acc$(\epsilon(K, N))$; $Q_{in}$.append(acc.)
            \IF {~$|Q_{in}|$ == $\gamma$ \textbf{and} div($Q_{in}$)$>\lambda_2$ \textbf{or} $l_{c}^{in}>l_{in}$ ~}
            \STATE $l_{c}^{in}$ = 0, $Q_{in}$ = [].
    		\STATE \textbf{Break}
    		\ELSE \STATE $Q_{in}$.pop[0].
    		\ENDIF
    		\ENDWHILE
    		\ENDWHILE
    		\ENDWHILE
	\end{algorithmic}
    \label{alg-opt}
\end{algorithm}

\subsection{Dynamic Bi-level Optimization}
Similar to previous work \cite{wang2018dataset, zhao2021DC}, we also learn the synthetic set with a bi-level optimization, in which the synthetic set $\ms$ is updated using Eq. \ref{eq.updateS} in the outer-loop and network parameters $\btheta$ is updated using 
\begin{equation}
    \btheta \leftarrow \argmin_{\btheta} J(\ms, \btheta)
\label{eq.updateTheta}
\end{equation}
in the inner-loop alternatively. $J(\ms, \btheta)$ calculates the cross-entropy classification loss on the synthetic set $\ms$. In this way, the synthetic set can be trained on many different $\btheta$ so that it can generalize to them. We initialize $\ms$ and $\btheta$ from random noise and standard network random initialization \cite{he2015delving}.
Previous work \cite{zhao2021DC, zhao2021DSA} sets a fixed number of outer-loop and inner-loop optimization steps, which takes too much time to adjust the hyper-parameters and may lead to networks' over- or under-fitting on synthetic set. To address these issues, we design a new bi-level optimization algorithm that can break the outer- and inner-loop automatically. Fig. \ref{pipeline} illustrates the proposed dynamic bi-level optimization module. To monitor the changing of network parameters $\btheta$, we randomly sample some images from real training set as a query set to evaluate the network. Then, a queue $Q$ is used to store the performance on the query set. We expect to learn synthetic data on more diverse network parameters. Hence, we sample inner-loop networks to optimize synthetic images when remarkable performance improvement is achieved on the query set. The optimization will be stopped when the performance on the query set is converged. $\lambda_1$ and $\lambda_2$ are two hyper parameters of dynamic bi-level optimization. 
We implement ablation study to show that the performance is not sensitive to $\lambda_1$ and $\lambda_2$.
The training algorithm is summarized in Alg. \ref{alg-opt}.

\begin{table*}[tp]
\vspace{-10pt}
\renewcommand\arraystretch{1.0}
\centering
\scriptsize
\setlength{\tabcolsep}{2pt}
\caption{The performance (testing accuracy \%) comparison to state-of-the-art methods.  LD$^{\dag}$ and DD$^{\dag}$ use LeNet for MNIST and AlexNet for CIFAR10, while the rest use ConvNet for training and testing. IPC: Images Per Class, Ratio~(\%): the ratio of condensed images to whole training set.}
\vspace{-5pt}
\begin{tabular}{ccccccccccccccc}
\toprule
\multirow{2}{*}{}           & \multirow{2}{*}{IPC} & \multirow{2}{*}{Ratio \%} & \multicolumn{4}{c}{Coreset Selection} 
&\multicolumn{6}{c}{Condensation}    & \multirow{2}{*}{Whole Dataset} \\ 
                            &                          &                           & Random          & Herding         & K-Center       & Forgetting &DD$^{\dag}$ &LD$^{\dag}$   &DC &DSA &CAFE        & CAFE+DSA      \\ \midrule
\multirow{3}{*}{MNIST}          & 1   & 0.017  & 64.9$\pm$3.5  & 89.2$\pm$1.6  & 89.3$\pm$1.5  & 35.5$\pm$5.6    &-&60.9$\pm$3.2& 91.7$\pm$0.5 &88.7$\pm$0.6 &\textbf{93.1$\pm$0.3}   &{90.8$\pm$0.5} & \multirow{3}{*}{99.6$\pm$0.0} \\ 
                                & 10  & 0.17   & 95.1$\pm$0.9  & 93.7$\pm$0.3  & 84.4$\pm$1.7  & 68.1$\pm$3.3   &79.5$\pm$8.1&87.3$\pm$0.7& 97.4$\pm$0.2 &\textbf{97.8$\pm$0.1} &97.2$\pm$0.2 &97.5$\pm$0.1  &\\ 
                                & 50  & 0.83   & 97.9$\pm$0.2  & 94.8$\pm$0.2  & 97.4$\pm$0.3  & 88.2$\pm$1.2   &-&93.3$\pm$0.3& 98.8$\pm$0.2 &\textbf{99.2$\pm$0.1} &98.6$\pm0.2$ &98.9$\pm0.2$ & \\ \midrule

\multirow{3}{*}{FashionMNIST}   & 1   & 0.017  & 51.4$\pm$3.8  & 67.0$\pm$1.9  & 66.9$\pm$1.8  & 42.0$\pm$5.5  &-&- & 70.5$\pm$0.6 &70.6$\pm$0.6 &\textbf{77.1$\pm$0.9} &{73.7$\pm$0.7} & \multirow{3}{*}{93.5$\pm$0.1} \\ 
                                & 10  & 0.17   & 73.8$\pm$0.7  & 71.1$\pm$0.7  & 54.7$\pm$1.5  & 53.9$\pm$2.0    &-&-& 82.3$\pm$0.4 &\textbf{84.6$\pm$0.3} &{83.0$\pm$0.4} &83.0$\pm$0.3  &         \\ 
                                & 50  & 0.83   & 82.5$\pm$0.7  & 71.9$\pm$0.8  & 68.3$\pm$0.8  & 55.0$\pm$1.1  &-&- & 83.6$\pm$0.4 &\textbf{88.7$\pm$0.2} &{84.8$\pm$0.4} &88.2$\pm$0.3 & \\ \midrule

\multirow{3}{*}{SVHN}           & 1   & 0.014  & 14.6$\pm$1.6  & 20.9$\pm$1.3  & 21.0$\pm$1.5  & 12.1$\pm$1.7  &-&-&  31.2$\pm$1.4 &27.5$\pm$1.4 &\textbf{42.6$\pm$3.3} &\textbf{42.9$\pm$3.0} & \multirow{3}{*}{95.4$\pm$0.1} \\ 
                                & 10  & 0.14   & 35.1$\pm$4.1  & 50.5$\pm$3.3  & 14.0$\pm$1.3  & 16.8$\pm$1.2   &-&-& 76.1$\pm$0.6 &\textbf{79.2$\pm$0.5} &75.9$\pm$0.6 & 77.9$\pm$0.6 &   &    \\
                                & 50  & 0.7    & 70.9$\pm$0.9  & 72.6$\pm$0.8  & 20.1$\pm$1.4  & 27.2$\pm$1.5   &-&-& 82.3$\pm$0.3 &\textbf{84.4$\pm$0.4} &{81.3$\pm$0.3} &82.3$\pm$0.4 & \\ \midrule 

\multirow{3}{*}{CIFAR10}        & 1   & 0.02   & 14.4$\pm$2.0  & 21.5$\pm$1.2  & 21.5$\pm$1.3  & 13.5$\pm$1.2   &-&25.7$\pm$0.7& 28.3$\pm$0.5 &28.8$\pm$0.7 &{30.3$\pm$1.1} &\textbf{31.6$\pm$0.8} & \multirow{3}{*}{84.8$\pm$0.1}         \\ 
                                & 10  & 0.2    & 26.0$\pm$1.2  & 31.6$\pm$0.7  & 14.7$\pm$0.9  & 23.3$\pm$1.0   &36.8$\pm$1.2&38.3$\pm$0.4& 44.9$\pm$0.5 &\textbf{52.1$\pm$0.5} &{46.3$\pm$0.6}  &50.9$\pm$0.5  &         \\ 
                                & 50  & 1      & 43.4$\pm$1.0  & 40.4$\pm$0.6  & 27.0$\pm$1.4  & 23.3$\pm$1.1   &-&42.5$\pm$0.4& 53.9$\pm$0.5 &60.6$\pm$0.5 &{55.5$\pm$0.6} &\textbf{62.3$\pm$0.4} & \\ \midrule
\multirow{3}{*}{CIFAR100}   & 1   & 0.2    &  4.2$\pm$0.3  &  8.4$\pm$0.3  &  8.3$\pm$0.3  &  4.5$\pm$0.3  &-&11.5$\pm$0.4& 12.8$\pm$0.3  & \textbf{13.9$\pm$0.3} &{12.9$\pm$0.3} &\textbf{14.0$\pm$0.3}  & \multirow{3}{*}{56.17$\pm$0.3}          \\
                            & 10  & 2      & 14.6$\pm$0.5  & 17.3$\pm$0.3  &  7.1$\pm$0.2  &  9.8$\pm$0.2  &-&-& 25.2$\pm$0.3  & \textbf{32.3$\pm$0.3} &{27.8$\pm$0.3} &31.5$\pm$0.2        \\
                             & 50  & 10      & 30.0$\pm$0.4  & 33.7$\pm$0.5  &  30.5$\pm$0.3  &  -  &-&-& -  & \textbf{42.8$\pm$0.4} & {37.9$\pm$0.3} &\textbf{42.9$\pm$0.2}        \\
                            \bottomrule
\end{tabular}

\label{tab:sota_coreset}
\end{table*}

\section{Experiments}
In this section, 
we first introduce the used datasets and implementation details. Then, we compare the proposed method to the state-of-the-art methods. After that, we conduct sufficient ablation studies to analyze the significant components and the influence of hyper parameters. Finally, the visualizations of synthetic images and feature distributions are provided to show the superiority of our CAFE.


\subsection{Datasets \& Implementation Details}
\textbf{MNIST \cite{lecun1998gradient}.}
The MNIST is a handwritten digits dataset that is commonly used for validating image recognition models. It contains 60,000 training images and 10,000 testing images with the size of 28$\times$28.

\textbf{FashionMNIST \cite{xiao2017fashion}.}
FashionMNIST is a dataset of Zalando's article images, consisting of a training set of 60,000 examples and a test set of 10,000 examples. Each example is a 28$\times$28 gray-scale image, associated with a label from 10 classes. 

\textbf{SVHN \cite{sermanet2012convolutional}.}
SVHN is a real-world image dataset for developing machine learning and object recognition algorithms. It consists of over 600,000 digit images coming from real world data. The images are cropped to 32$\times$32.

\textbf{CIFAR10/100 \cite{krizhevsky2009learning}.} The two CIFAR datasets consist of tiny colored natural images with the size of 32$\times$32 from 10 and 100 categories, respectively. In each dataset, 50,000 images are used for training and 10,000 images for testing.

\paragraph{Implementation Details.}
We present the experiments details of the outer-loop and inner-loop, respectively. In outer-loop, we optimize 1/10/50 Images Per Class (IPC) synthetic sets for all the five datasets using three-layer Convolutional Network (ConvNet) as same as \cite{zhao2021DC}. The ConvNet includes three repeated ``Conv-InstNorm-ReLU-AvgPool" blocks. The channel number of each convolutional layer is 128. The initial learning rate of synthetic images is 0.1, which is divided by 2 in 1,200, 1,400, and 1,800 iterations. We stop training in 2,000 iterations. For inner-loop, we train the ConvNet on synthetic sets  for 300 epochs and evaluate the performances on 20 randomly initialized networks. The initial learning rate of network is 0.01. Following \cite{zhao2021DC}, we perform 5 experiments and report the mean and standard deviation on 100 networks. The default $N$ is 256, $\lambda_1$ is 0.05 and $\lambda_2$ is 0.05. We assess the sensitiveness of $\lambda_1$ and $\lambda_2$ in the Sec. \ref{abl_stdy}.

\subsection{Comparison to the State-of-the-art Methods}
We compare our method to four coreset selection methods, namely Random \cite{chen2010super, rebuffi2017icarl}, Herding \cite{castro2018end, belouadah2020scail}, K-Center \cite{farahani2009facility, sener2017active} and Forgetting \cite{toneva2019empirical}. We also make comparisons to recent state-of-the-art condensation methods, namely Dataset Distillation (DD) \cite{wang2018dataset}, LD \cite{bohdal2020flexible}, Dataset Condensation (DC) \cite{zhao2021DC} and DSA (adding differentiable Siamese augmentation for DC) \cite{zhao2021DSA}. \bo{Although state-of-the-art performances are achieved in \cite{nguyen2021dataset, nguyen2021datasetnips}, we do not compare to them due to the remarkable difference in training scheme and cost.} We report the performances of our method and competitors on five datasets in Tab. \ref{tab:sota_coreset}. When learning 1 image per class, our method achieves the best results on all the 5 datasets. In particular, the improvements on SVHN and FashionMNIST are 11\% and 6.5\% over other methods. 
Condensation-based methods outperforms than coreset selection methods with a large margin.
Among coreset selection methods, Herding and K-Center outperform Random and Forgetting with a large margin.
When learning 10 and 50 images/class, the performance of our method exceeds DC with 0.7\%$\sim$2.6\% on most datasets. Compared with DSA, 
Our CAFE+DSA achieves comparable results with DSA on most datasets on CIFAR10/100. For 50 images/class learning on CIFAR10, our CAFE+DSA outperforms DSA by 1.7\%.


\begin{table}[]
\vspace{-10pt}
\centering
\small
\caption{Evaluation of the three components in CAFE}
\vspace{-5pt}
\label{tab:2}
\begin{tabular}{cccccc|ccc|}
	\toprule
	DL & LFA &Dynamic Bi-level Opt.  & Performance     \\
	\midrule
	\checkmark &  &   &  49.78 \\
	 &\checkmark  &   &53.96   \\
	\checkmark &\checkmark  &   &  54.53 \\
	\checkmark &  &\checkmark   &  50.92 \\
	 &\checkmark  &\checkmark   &  54.98 \\
	\checkmark &\checkmark  & \checkmark  & \bf 55.50 \\
\bottomrule
\end{tabular}
\end{table}

\subsection{Ablation Studies}
\label{abl_stdy}
In this subsection, we study ablations using CIFAR10 (IPC = 50) to investigate the effectiveness of each module and the influence of the hyper parameters.

\paragraph{Evaluation of the three components in CAFE.}
To explore the effect of each component in our method, we design ablation studies of Discrimination Loss (DL), Layer-wise Feature Alignment (LFA) and Dynamic Bi-level Optimization on CIFAR10. As shown in Tab. \ref{tab:2}, DL, LFA and Dynamic Bi-level Opt. are complementary with each other. CAFE performs poorly when using DL individually (49.78\%), as DL focuses more on classifying the real samples but ignores the distribution consistency with real images.
The result of using LFA individually outperforms DL with 4.18\%, which implies considering of distribution consistency is more important for dataset condensation. However, utilizing LFA independently means the importance of all the images in real dataset are equal, which may overlook the information from discriminative samples (\textit{i.e.} samples nearby the decision boundaries). Jointly using DL and LFA can obtain better result than using DC on CIFAR10 testing set. Adding the Dynamic Bi-level Opt. can further improve the performance of DL and LFA, which indicates breaking out-/inner-looper automatically can reduce the over-/under-fitting effectively. 
Using these three components together achieves the highest result. To understand the effect of DL and LFA more intuitively, we also visualize the synthetic images feature distributions of using DL or LFA independently in Sec. \ref{visualization}.

\begin{table}[]
\vspace{-10pt}
\small
\centering
\caption{Evaluation of the importance of layer-wise feature alignment. The layer1 is closest to the output layer while layer4 is closest to input layer. Note that, layer4 represents the last average pooling layer in ConvNet.}
\vspace{-5pt}
\label{tab:3}
\begin{tabular}{cccccc|ccc|}
	\toprule
	Layer1 & Layer2 &Layer3.  & Layer4 & Performance/+DL     \\
	\midrule
	\checkmark &  &   & & 50.74/\textbf{52.78} \\
	 &\checkmark  &   & &  43.45/49.30 \\
	 &  &\checkmark   & &  44.52/49.08 \\
	 &  &   &\checkmark & \textbf{51.30}/52.05 \\

\bottomrule
\end{tabular}
\end{table}

\paragraph{Exploring the importance of layer-wise feature alignment in each layer.}
To investigate the importance of feature alignment, we apply the feature alignment operation to each layer individually. As shown in Tab. \ref{tab:3}, the performances of different layers vary remarkably. Applying feature alignment operation in layer1 or layer4 obtains better results than in layer2 or layer3, as the supervision in layer2 or layer3 is far from the input and output layers.
Applying feature alignment in each layer individually can not obtain promising results. To demonstrate the effectiveness of DL in each layer, we also show the results of adding DL loss. The addition of DL can consistently improve the performances in all layers.

\begin{table}[]
\vspace{-5pt}
\small
\centering
\caption{Evaluation of complementarity of layer-wise feature alignment. The indexes of layers are same as Tab. \ref{tab:3}.}
\vspace{-5pt}
\label{tab:4}
\begin{tabular}{cccccc|ccc|}
	\toprule
	Layer1 & Layer2 &Layer3.  & Layer4 & Performance/+DL     \\
	\midrule
	\checkmark &  &   & & 50.74/52.78 \\
	\checkmark &\checkmark  &   & & 51.27/53.28 \\
	\checkmark &\checkmark  &\checkmark   & & 53.16/53.96 \\
	\checkmark &\checkmark  &\checkmark   &\checkmark & \textbf{54.98}/\textbf{55.50} \\

\bottomrule
\end{tabular}
\end{table}

\begin{figure*}[tp]
\vspace{-10pt}
    \centering
    \begin{subfigure}{0.33\textwidth}
        \includegraphics[width=\linewidth]{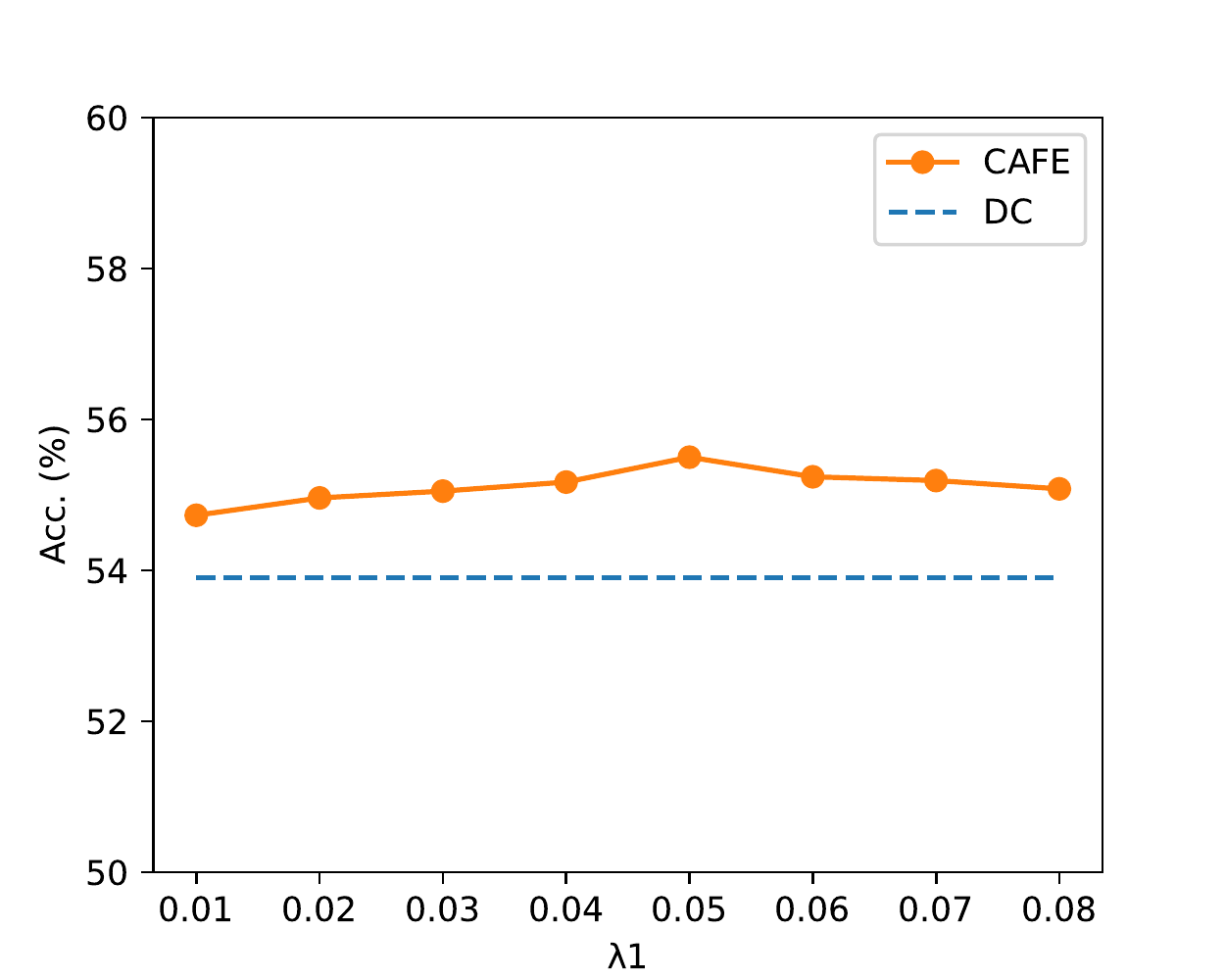}
        \caption{Ablation of $\lambda_1$.}
        \label{lambda1}
    \end{subfigure}
    \begin{subfigure}{0.33\textwidth}
        \includegraphics[width=\linewidth]{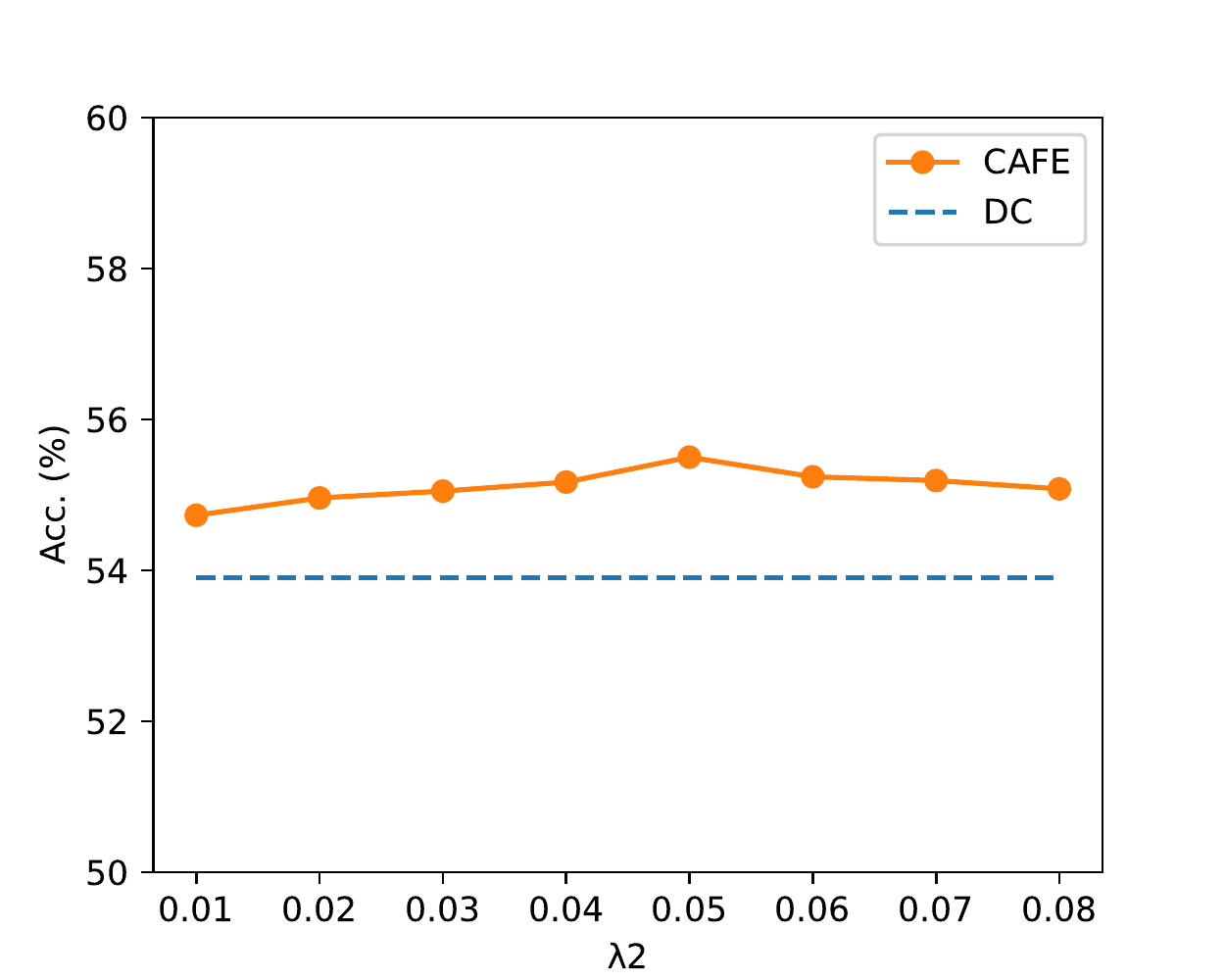}
        \caption{Ablation of $\lambda_2$.}
        \label{lambda2}
    \end{subfigure}
    \begin{subfigure}{0.33\textwidth}
        \raisebox{-4.4cm}{\includegraphics[width=\linewidth, height=0.7\linewidth]{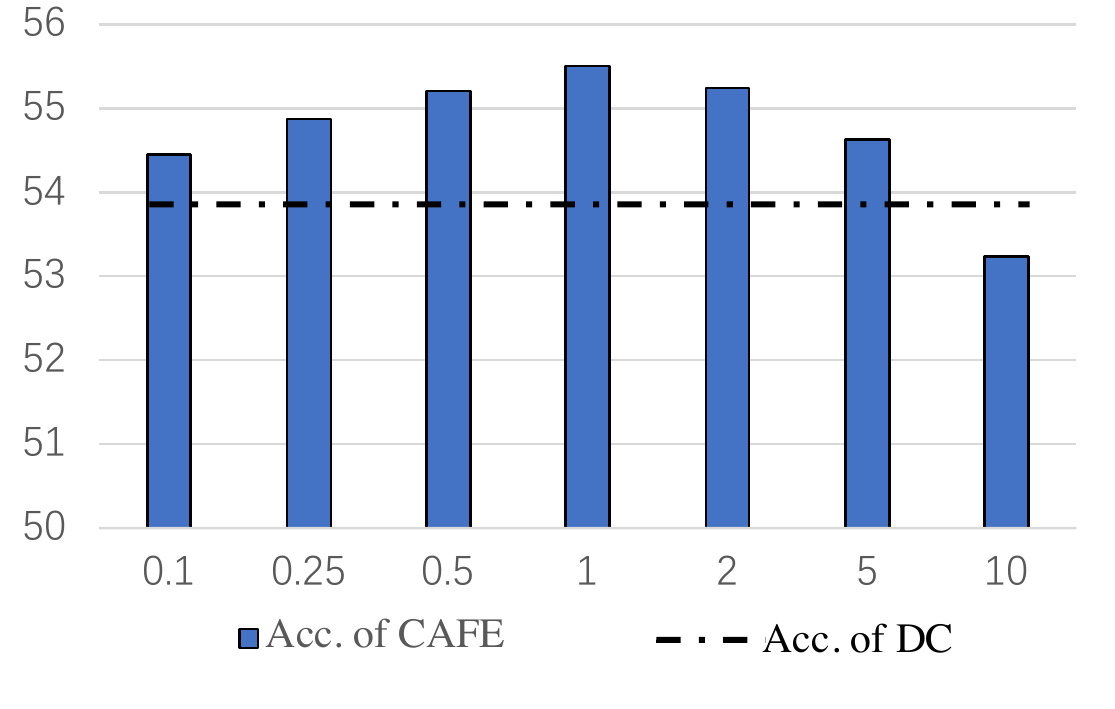}}
        \caption{Ablation of $\beta$.}
        \label{beta}
    \end{subfigure}
\vspace{-5pt}
\caption{
$\lambda_1$ and $\lambda_2$ are the hyper-parameters in dynamic bi-level optimization module. $\beta$ is the ratio between $\mathcal{L}_{\text{f}}$ and $\mathcal{L}_{\text{d}}$.}
\label{fig:k_alpha}
\end{figure*}

\paragraph{Exploring the complementarity of layer-wise feature alignment among all layers.}
After evaluating the importance of the LFA in each layer, exploring the complementarity of LFA in all layers is also very important.
 We first utilize the feature alignment in the layer1 to update the synthetic images. Then, we apply the same feature alignment to other layers (\textit{i.e.} layer2, layer3, layer4). Here, we also consider the effect of DL and report the results with and without using DL in Tab. \ref{tab:4}. When adding feature alignment to more layers, the performances on testing set become better. Meanwhile,
the DL can further improve the performance in all cases. Specifically, the performance difference between using LFA in all layers and using in the first layer is about 4\% (w/o DL) and 3\% (with DL). The average performance boost of adding each layer is about 1\% (w/o DL) and 0.7\% (with DL), which indicates the strong complementarity of layer-wise feature alignment in all layers. 

\begin{table}[]
\vspace{-10pt}
\centering
\scriptsize
\caption{Evaluation of $\gamma$ and the training time.}
\vspace{-5pt}
\label{tab:5}
\begin{tabular}{ccccc|cccc|}
	\toprule
	$\gamma$ & 5 &10  & 15 & 20  &DC    \\
	\midrule
	Accuracy (\%) &53.16 &\textbf{55.50} &54.10&53.63&53.9\\
	Time (minutes) &$ \approx$117 &$ \approx$367 & $ \approx$463 & $ \approx$463 &$ \approx$460\\

\bottomrule
\end{tabular}
\end{table}

\paragraph{Evaluation of $\lambda_1$ and $\lambda_2$.}
$\lambda_1$ and $\lambda_2$ are the thresholds to control whether break the out-looper and inner-looper or not. As shown in Fig. \ref{lambda1} and \ref{lambda2}, we study different values of $\lambda_1$ and $\lambda_2$ ranging from 0.01 to 0.08. Our default $\lambda_1$ = 0.05 and $\lambda_2$ = 0.05 achieve the best results, which outperforms DC 1.6\%. For out-looper, too large $\lambda_1$ may reduce the iterations of updating the synthetic images, which leads to the worse results. 
Too small $\lambda_1$ increases the optimization difficulties and even makes the model unable to break the out-looper normally. As for inner-looper, the model diversity is not large enough when $\lambda_2$ is too small, whereas it would be tricky to break the inner-looper when $\lambda_2$ is very large. Furthermore, it is worth noting that our method outperforms DC with a large margin at almost all settings. Meanwhile, the performance is not sensitive to $\lambda_1$ and $\lambda_2$. 

\paragraph{Evaluation of the ratio $\beta$.}
In Fig. \ref{beta}, we evaluate the effect of different ratios between the $\mathcal{L}_{\text{f}}$ and $\mathcal{L}_{\text{d}}$. We find that setting equal weight for each loss achieves the best results. The performance gets promoted as $\beta$ increase from 0.1 to 1, while increasing the weight of $\mathcal{L}_{\text{d}}$ from 1 to 10 dramatically degrades performance. 

\paragraph{Evaluation of $\gamma$ and the training time.}
$\gamma$ is hyper parameter of the maximum length of queues in dynamic bi-level optimization. We show the performances and training time of different $\gamma$ in Tab. \ref{tab:5}. One can find the default $\gamma = 10$ achieves the best result and requires less time than \textit{DC}. Too small and too large $\gamma$ may lead to under- or over-fitting.

\paragraph{Evaluation of the generalization to unseen architectures}
To evaluate the generalization ability of synthetic data on unseen architectures, we first condense CIFAR10 dataset with ConvNet to generate synthetic images. Then, we train different architectures, including AlexNet, VGG11, ResNet18, and MLP  (3 layers), on the synthetic images. As shown in Tab. \ref{tab:cross-arch}, our method achieves better generalization performance than DC obviously. Specifically, our method outperforms DC with 5.25\%, 1.79\%, 4.42\%, and 7.96\% when testing on AlexNet, VGG11, ResNet18, and MLP (3 layers). 

\begin{table}[]
\vspace{-10pt}
\centering
\setlength{\tabcolsep}{2pt}
\scriptsize
\caption{The testing performance (\%) on unseen architectures. The 50 IPC synthetic set is learned on one architecture (C), and then tested on another architecture (T).}
\vspace{-5pt}
\label{tab:cross-arch}

\begin{tabular}{ccccccc}
\toprule
&   \texttt{C}\textbackslash \texttt{T} & ConvNet      & AlexNet      & VGG11          & ResNet18  &MLP           \\ 	\midrule
DC                 & ConvNet           & 53.9$\pm$0.5 & 28.77$\pm$0.7 & 38.76$\pm$1.1 & 20.85$\pm$1.0  &28.71$\pm$0.7 \\
\multirow{1}{*}{CAFE} & ConvNet           & \textbf{55.50$\pm$0.4} & \textbf{34.02$\pm$0.6} & \textbf{40.55$\pm$0.8} & \textbf{25.27$\pm$0.9} & \textbf{36.67$\pm$0.6} \\
                    \bottomrule
\end{tabular}

\end{table}




\begin{figure*}[t]
\vspace{-10pt}
\centering
\subfloat[Original CIFAR10 images.]{\includegraphics[width = .32\linewidth]{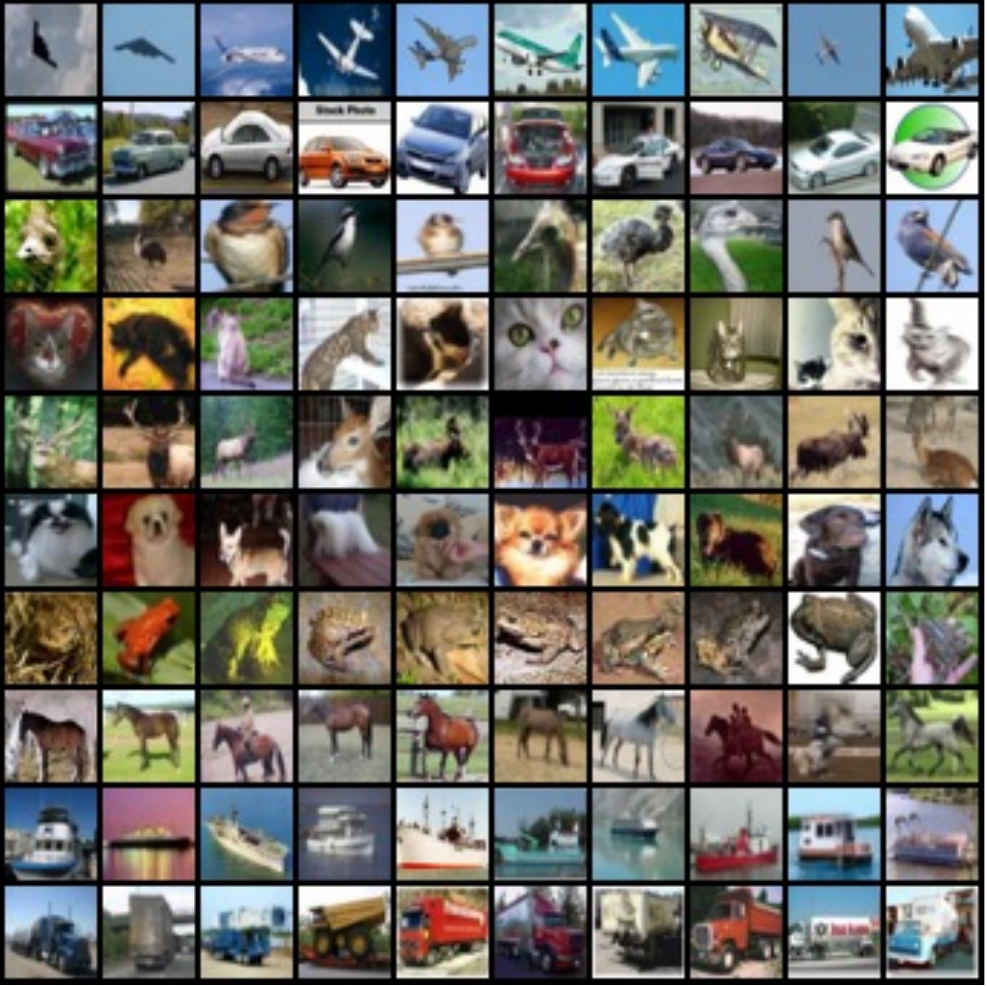}\label{fig:vis_a}}\hfill
\subfloat[The synthetic images of \textbf{CAFE}.]{\includegraphics[width = .32\linewidth]{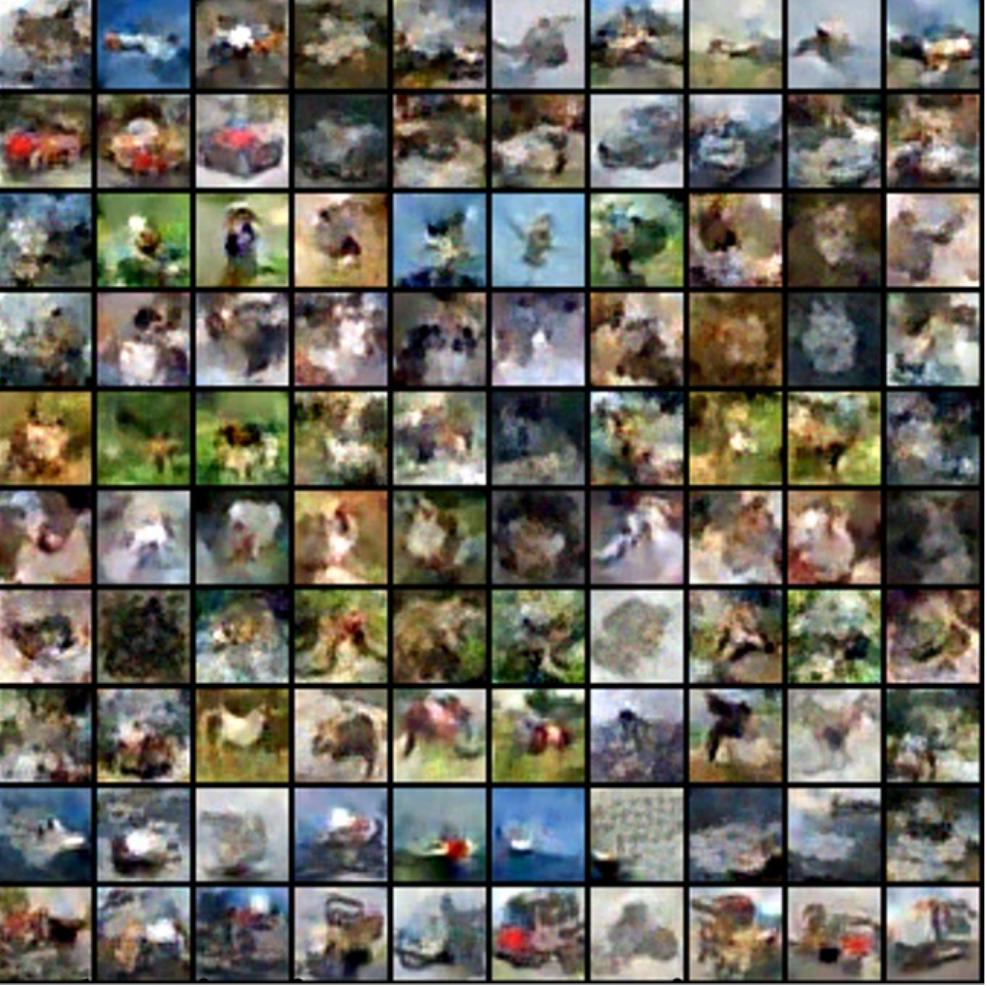}\label{fig:vis_b}}\hfill
\subfloat[The synthetic images of DC.]{\includegraphics[width = .32\linewidth]{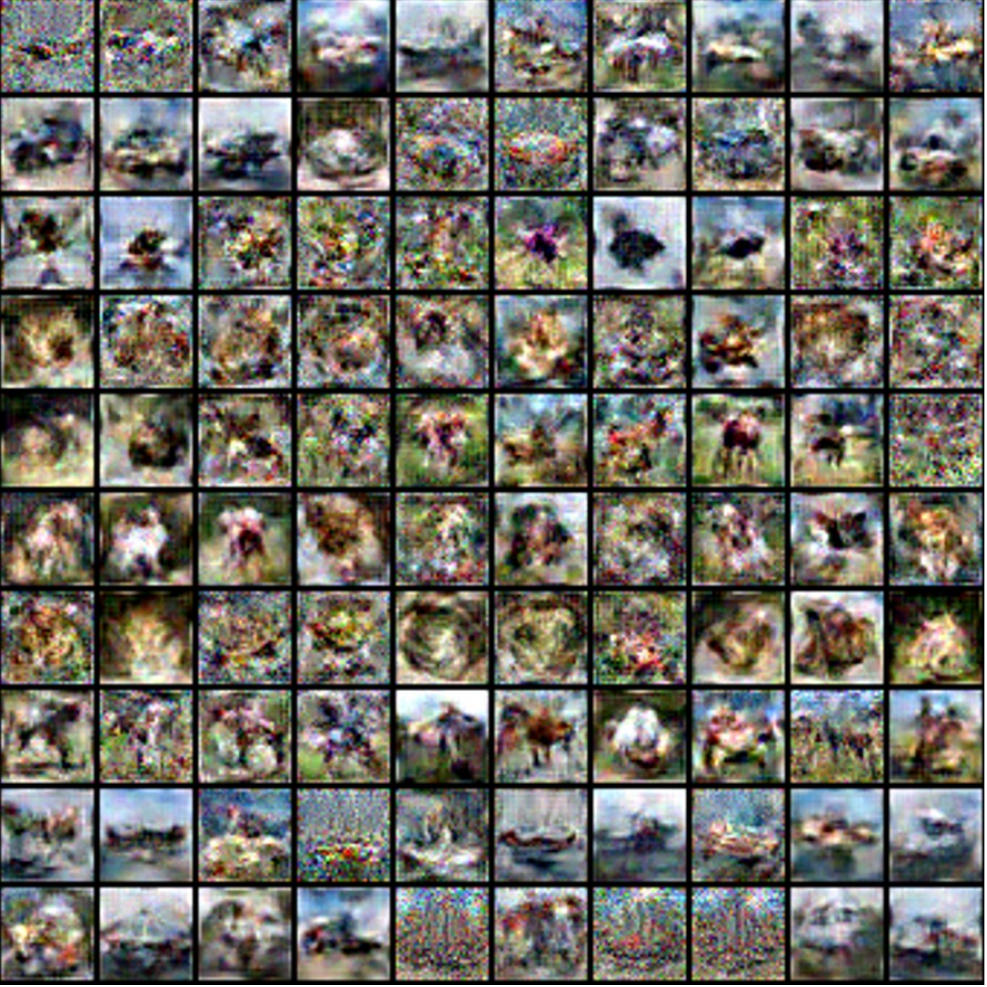}\label{fig:vis_c}}
\vspace{-5pt}
\caption{Visualizations of original images, and synthetic images generated by CAFE and DC. Both CAFE and DC are initialized from random noise.}     
\label{vis}
\end{figure*}
\subsection{Visualizations}
In this subsection, we visualize the synthetic images as well as data distribution to show the effectiveness of CAFE.
\label{visualization}


\paragraph{Synthetic images.}
To make fair comparison, the synthetic set is initialized by the same random noise (IPC = 50). After that, we apply DC and CAFE to optimize the synthetic set on CIFAR10 dataset. Finally, the partial (only show 10 images per class) optimized synthetic images and original images of CIFAR10 are shown in Fig. \ref{vis}. There are several observations can be summarized as follows: 1). It is easy to find that the synthetic images generated by our method is more visually similar to original CIFAR10 images than DC. 2). The synthetic images have more semantic information than DC, which illustrates the effectiveness of LFA and DL modules. 3). A certain ratio of images generated by DC are not very clear, which could not provide enough discriminative features for classification. 

\begin{figure}[t]
\vspace{-5pt}
\centering
\includegraphics[width=0.45\textwidth]{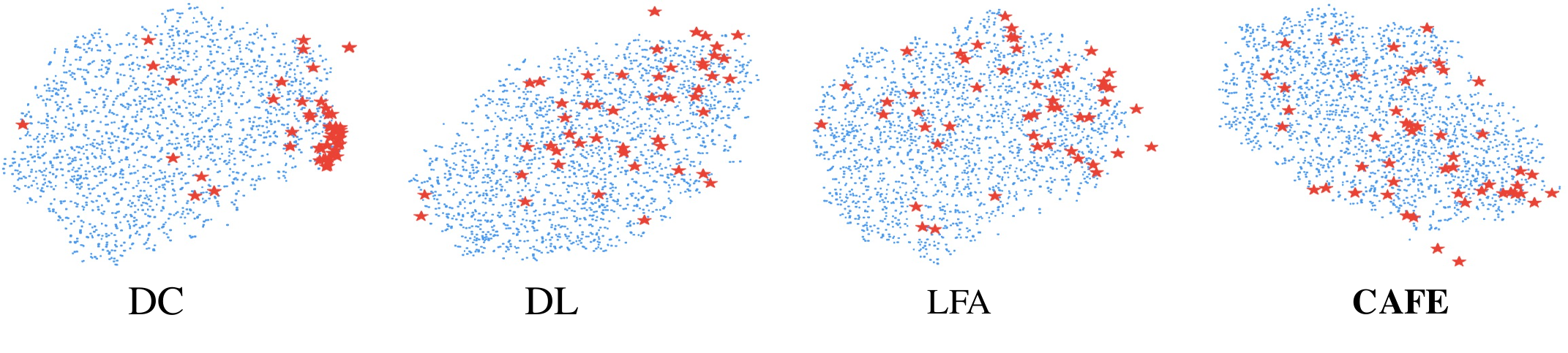}
\vspace{-5pt}
\caption{The data distribution of real images and synthetic images learned by DC \cite{zhao2021DC}, DL, LFA, and CAFE for one category in CIFAR10.}
\label{fig:distribution}
\vspace{-5pt}
\end{figure}

\paragraph{Data distribution.}
To evaluate whether the synthetic images using our method can capture more accurate distribution from original dataset, we utilize t-SNE to visualize the features of real set and synthetic sets generated by DC, DL, LFA and CAFE. As shown in Fig. \ref{fig:distribution}, the ``points" and ``stars" represent the real and synthetic features. The synthetic images of DC gather around a small area of decision boundary, which indicates using DC can not capture the original distribution well. Our methods DL, LFA, and CAFE effectively capture useful information across the whole real dataset, which possesses a good generalization among different CNN architectures.





\section{Conclusion}
In this work, we propose a novel scheme to Condense dataset by Aligning FEatures (CAFE), which explicitly attempts to preserve the real-feature distribution as well as the discriminant power of the resulting synthetic data, lending itself to strong generalization capability to unseen architectures. The CAFE consists of three carefully designed modules, namely layer-wise feature alignment module, discrimination loss, and dynamic bi-level optimization module. The feature alignment module and discrimination loss concern capturing distribution consistency between synthetic and real sets,
while bi-level optimization enables CAFE to 
learn customized SGD steps to avoid over-/under-fitting.
Experimental results across
various datasets demonstrate
that, CAFE consistently outperforms the state of the art
with less computation cost, 
making it 
readily applicable to in-the-wild scenarios.
As the future work, we plan to explore the use of dataset condensation on more challenging datasets such as ImageNet \cite{deng2009imagenet}.

\textbf{Acknowledge.} This research is supported by the National Research Foundation, Singapore under its AI Singapore Programme (AISG Award No: AISG2-PhD-2021-08-008), NUS ARTIC Project (ECT-RP2), China Scholarship Council 201806010331 and the EPSRC programme grant Visual AI EP/T028572/1. We thank Google TFRC for supporting us to get access to the Cloud TPUs. We thank CSCS (Swiss National Supercomputing Centre) for supporting us to get access to the Piz Daint supercomputer. We thank TACC (Texas Advanced Computing Center) for supporting us to get access to the Longhorn supercomputer and the Frontera supercomputer. We thank LuxProvide (Luxembourg national supercomputer HPC organization) for supporting us to get access to the MeluXina supercomputer.




{\small
\bibliographystyle{ieee_fullname}
\bibliography{refs}
}

\end{document}